\documentclass[sigconf]{acmart}
\AtBeginDocument{%
  }

\copyrightyear{2025}
\acmYear{2025}
\setcopyright{acmlicensed}\acmConference[ICMR '25]{Proceedings of the 2025 International Conference on Multimedia Retrieval}{June 30-July 3, 2025}{Chicago, IL, USA}
\acmBooktitle{Proceedings of the 2025 International Conference on Multimedia Retrieval (ICMR '25), June 30-July 3, 2025, Chicago, IL, USA}
\acmDOI{10.1145/3731715.3733326}
\acmISBN{979-8-4007-1877-9/2025/06}
\usepackage{multicol, multirow}
\usepackage{amsmath}

\usepackage{amssymb, amsthm}
\usepackage{float}
\usepackage{mathtools}
\usepackage{array}
\usepackage{float}
\usepackage{bbding}
\usepackage{units}
\usepackage{makecell}
\usepackage[linewidth=1pt]{mdframed}

\usepackage{amsmath,amsfonts,amsthm}
\usepackage{amssymb}
\usepackage{mathtools}
\usepackage{array}

\newtheorem{theorem}{Theorem}
\usepackage{algorithm}

\usepackage{libertine}
\begin{document}

\title{Enhancing OOD Detection Using Latent Diffusion}

\author{Heng Gao}
\email{hgao22@m.fudan.edu.cn}
\affiliation{%
  \institution{Fudan University}
  \city{Shanghai}
  \country{China}
}

\author{Jun Li}
\authornote{Corresponding author.}
\email{jun_li@fudan.edu.cn}
\affiliation{%
  \institution{Fudan University}
  \city{Shanghai}
  \country{China}
}









\begin{abstract}
    Out-of-distribution (OOD) detection is crucial for the reliable deployment of machine learning models in real-world scenarios, enabling the identification of unknown samples or objects.  A prominent approach to enhance OOD detection performance involves leveraging auxiliary datasets for training. Recent efforts have explored using generative models, such as Stable Diffusion (SD), to synthesize outlier data in the pixel space. However, synthesizing OOD data in the pixel space can lead to reduced robustness due to over-generation. To address this challenge, we propose Outlier-Aware Learning (OAL), a novel framework that generates synthetic OOD training data within the latent space, taking a further step to study how to utilize Stable Diffusion for developing a latent-based outlier synthesis approach. This improvement facilitates network training with fewer outliers and less computational cost. Besides, to regularize the model's decision boundary, we develop a mutual information-based contrastive learning module (MICL) that amplifies the distinction between In-Distribution (ID) and collected OOD data. Moreover, we develop a knowledge distillation module to prevent the degradation of ID classification accuracy when training with OOD data. The superior performance of our method on several benchmark datasets demonstrates its efficiency and effectiveness. Source code is available in \url{https://github.com/HengGao12/OAL}.
\end{abstract}

%
%





\maketitle
~

\begin{figure}[t]
 
\centerline{\includegraphics[width=\linewidth]{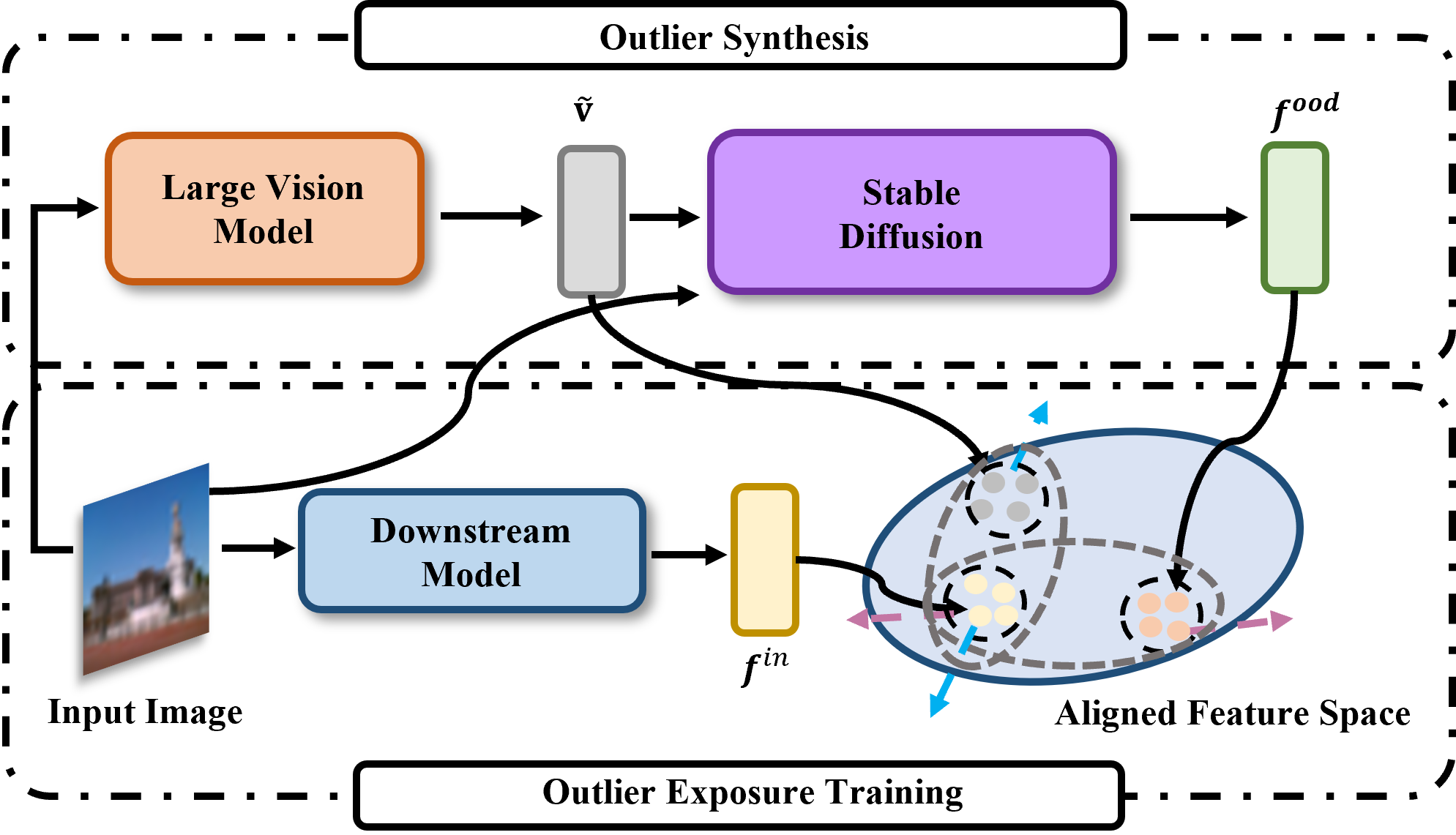}}
\caption{The illustration of the intuition behind the contrastive learning module MICL in OAL framework. Here, $\Tilde{\mathbf{v}}$ represents the OOD embeddings sampled from a vision model's feature space by k-Nearest Neighbor ($k$-NN) \cite{kramer2013k}, $f^{ood}$ denotes OOD latent embeddings transformed from the latent representation generated by Diffusion model, $f^{in}$ denotes the in-distribution feature of the downstream model. We enlarge the discrepancy between $f^{in}$, $f^{ood}$ and $\Tilde{\mathbf{v}}$ using MICL, thereby effectively regularizing the model's decision boundary.}
\Description{1}
\label{intuition}
\end{figure}

\section{Introduction}
Out-of-Distribution (OOD) detection is a crucial component for deploying models in the real world \cite{yang2024generalized,he2022ronf,chen2024fodfom}, which has drawn increasing attention in the machine learning safety community \cite{ming2022poem}. To prevent modern neural networks from being over-confident for OOD inputs and producing untrustworthy predictions \cite{nguyen2015deep,Fan_2024_CVPR}, several studies proposed training the model with an auxiliary dataset \cite{hendrycks2018deep} and regularizing the model to learn a compact decision boundary between in-distribution and out-of-distribution data \cite{du2022towards, NEURIPS2023_bf5311df}. These methods demonstrate superior OOD detection performance compared with those that do not use auxiliary datasets for training \cite{du2022towards, NEURIPS2023_bf5311df}. The main challenges for these approaches include: (i) how to collect representative OOD samples for outlier exposure training, and (ii) developing algorithms to effectively regularize the model using the generated OOD data.

\begin{figure}[t]
 
\centerline{\includegraphics[width=\linewidth]{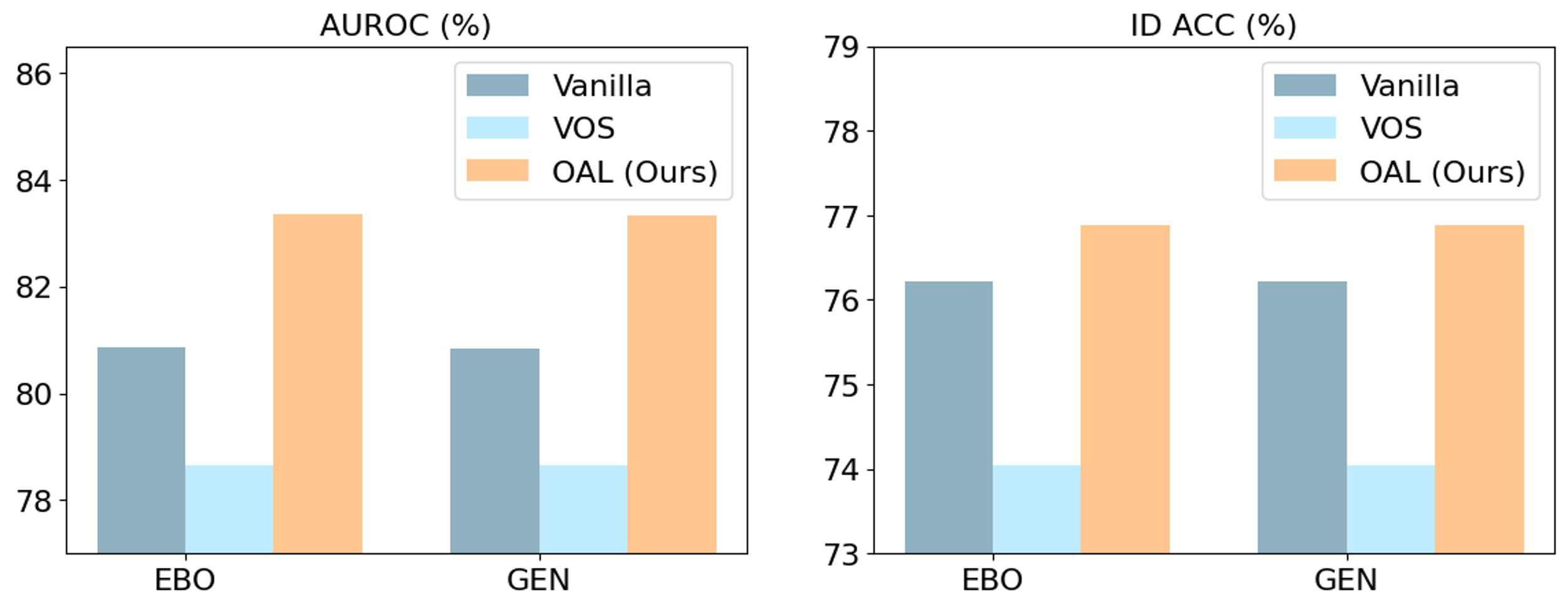}}
\caption{Performance comparisons. Here we compare the mean AUROC and ID accuracy of EBO \cite{liu2020energy} and GEN \cite{liu2023gen} scores on the CIFAR-100 benchmark using the vanilla training, the VOS and OAL training (Ours), respectively. From the above figure, we can find that VOS training may lead to ID performance (ID ACC) degradation while ours can improve both ID and OOD (AUROC) results of the model. For instance, OAL improves EBO score by $2.51\%$ in AUROC and $0.67\%$ in ID ACC. 
}
\Description{2}
\label{id-deg}
\end{figure}

In recent work, VOS \cite{du2022towards} proposes a latent-based approach that synthesizes outliers in the low-density region of the In-Distribution (ID) feature space and uses energy-based loss functions to regularize the model. In NPOS \cite{tao2023nonparametric}, the authors also propose a latent-based synthesis framework. Unlike VOS, they mitigate the issue of the unduly strong assumption on the model's ID feature space by using non-parametric nearest neighbor distance-based methods. Despite their effectiveness, these approaches may suffer from ID performance degradation after outlier exposure training (see experiment results in Figure \ref{id-deg}). 

To generate photo-realistic OOD data, enabling us to understand the synthesized outliers in a human-compatible manner, Dream-OOD \cite{NEURIPS2023_bf5311df} first propose to leverage Stable Diffusion \cite{rombach2022high} model to synthesize outlier images for training in the pixel space, marking a milestone in OOD detection. However, synthesizing images in the high dimensional pixel space can diminish the model robustness due to over-generation, i.e. generating too many similar or redundant samples \cite{narayanaswamy2023exploring}.  
Furthermore, we find that in Dream-OOD \cite{NEURIPS2023_bf5311df}, the ID accuracy also decreases after introducing the generated OOD data for training, even though the mean AUROC is improved.
However, the process of outlier exposure training must not harm the in-distribution classification accuracy \cite{yang2024generalized}. 

Motivated by these findings, we develop an \textbf{O}utlier \textbf{A}ware \textbf{L}earning (OAL) framework, aiming to take a further step to study how to leverage Stable Diffusion to develop latent-based synthesis approaches that generate outlier data in the feature space for facilitating outlier exposure training. We also develop a series of regularization algorithms to improve OOD detection performance without compromising the model's ID classification accuracy. 

To be specific, in our learning framework, to avoid making any assumptions on the model's ID feature space, we still utilize the k-Nearest Neighbor \cite{kramer2013k} method to sample outliers from the penultimate layer of a vision model. Then, to reduce the overall training time, we directly sample these outliers from a pre-trained Transformer \cite{dosovitskiy2020image} encoder, which has already achieved superior performance on the in-distribution classification task. Afterward, we take these $k$-NN sampled outliers as token conditions for Stable Diffusion to generate diverse OOD data in the latent space for supervision, which may reduce the worst-case performance of OE training theoretically (see Theorem 3.3 in \cite{mirzaeirodeo}). 

For regularizing the model, we develop a \textbf{M}utual \textbf{I}nformation-based \textbf{C}ontrastive \textbf{L}earning module (MICL) to enlarges the discrepancy between the ID and OOD data in the latent space, which is much better than the energy-based regularization methods \cite{du2022towards} (this result is validated in our ablation study). We treat the mutual information between ID and OOD samples as a signal to increase the distance between them in an aligned feature space, making the ID features and OOD features increasingly irrelevant.

Moreover, to further mitigate the model's overconfidence on OOD data \cite{nguyen2015deep,li2019reducing} while maintaining its inference speed \cite{vengertsev2023confidence}, we introduce a simple yet effective knowledge distillation module, named as IDKD (\textbf{I}n-\textbf{D}istribution \textbf{K}nowledge \textbf{D}istillation), preventing the ID performance degradation of the model. The experiments on CIFAR-10 and CIFAR-100 \cite{krizhevsky2009learning} benchmarks achieve non-trivial improvement in OOD detection performance, compared to both the post hoc methods and other outlier synthesis-based methods. 

The main contributions of our paper can be summarized as follows: 
\begin{itemize}
    \item We propose an Outlier Aware Learning (OAL) framework, which generates OOD data from the latent space of Stable Diffusion, thereby facilitating network training. Using pre-training techniques, our framework reduces about $\nicefrac{2}{3}$ training time overall compared with Dream-OOD. Besides, we find that synthesizing virtual outliers in the feature space requires much less storage resource consumption than synthesizing outliers in the pixel space.
    \item We develop a mutual information-based contrastive learning module (MICL) to better regularize the model's decision boundary between the ID and OOD data. Additionally, we propose a knowledge distillation module named IDKD to prevent ID accuracy degradation after outlier exposure training while maintaining the model's inference speed.
    \item We conduct extensive experiments on the CIFAR-10 and CIFAR-100 benchmarks. The results demonstrate that our approach achieves superior performance on various challenging OOD datasets, verifying the effectiveness of our diffusion-based outlier synthesis method.
\end{itemize}

\begin{figure*}[t]
 
\centerline{\includegraphics[width=0.98\linewidth]{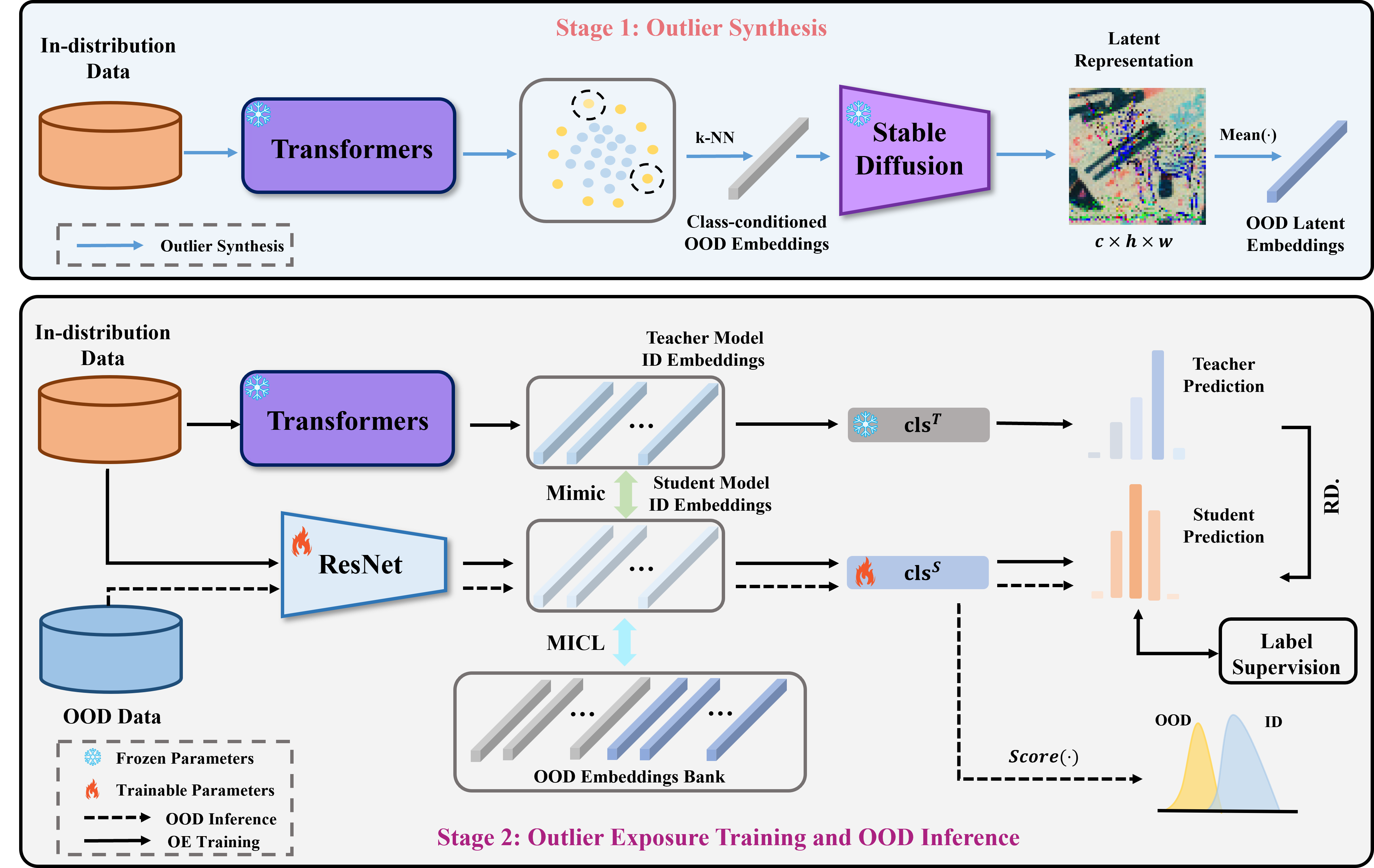}}
\caption{The overview of our OAL training pipeline for enhancing OOD detection. In stage 1, we first utilize $k$-NN distance search and Stable Diffusion to synthesize outliers in the latent space. Then, in stage 2, we develop a series of learning modules to regularize the model's decision boundary and boost its OOD detection performance by using the synthesized outliers. Note that, in the figure above, MICL represents the Mutual Information-based Contrastive Learning module. \lq RD.\rq~denotes response distillation. $\text{\textbf{\textit{Score}}}\left(\cdot\right)$ denotes the score function used in the OOD inference phase. Besides, in OAL, the score function used for detecting OOD samples is taken as EBO \cite{liu2020energy}.
}
\Description{3}
\label{OAL-pipeline}
\end{figure*}

\section{Related Work}
\noindent\textbf{Outlier exposure-based OOD detection.}\quad
The outlier exposure-based approaches leverage auxiliary datasets to train the model, thereby improving the model’s OOD detection performance \cite{hendrycks2018deep}. In VOS \cite{du2022towards}, Du et al. propose to synthesize OOD data automatically from a class-conditional distribution in the model's ID feature space and develop an energy-based loss function to regularize the model. In NPOS \cite{tao2023nonparametric}, the authors argue that the class-conditional Gaussian distribution assumption for the model's ID feature space is too restrictive and may not hold for more complex situations in the open world. Therefore, they propose to use a non-parametric k-Nearest Neighbor distance-based approach to sample outliers from the model's ID feature space. In Dream-OOD \cite{NEURIPS2023_bf5311df}, the authors first propose a Stable Diffusion-based \cite{rombach2022high} approach to synthesize photo-realistic OOD images which is easier for humans to understand than those latent-based synthesis approaches \cite{du2022towards, tao2023nonparametric}. 

In our study, different from Dream-OOD \cite{NEURIPS2023_bf5311df}, we synthesize outliers from the latent space of the Stable Diffusion model, which facilitates the model's training. 

\noindent\textbf{Representation learning-based OOD detection.}\quad
Another line of work leveraged representation learning techniques \cite{pmlr-v119-chen20j} to enhance the model's OOD detection performance. Sehwag et al. \cite{sehwag2021ssd} and Winkens et al. \cite{winkens2020contrastive} empirically demonstrate the effectiveness of representation learning approaches in OOD detection through contrastive learning methods. CSI \cite{tack2020csi} contrasts samples with distributionally shifted augmentations of themselves and introduces a new detection score adapted to this learning framework. SIREN \cite{du2022siren} develop a trainable loss function to shape the representation into a mixture of vMF distribution \cite{mardia2009directional} on the unit hyper-sphere and propose a new OOD score based on the learned class-conditional vMF distributions. 

In our research, we investigate the application of knowledge distillation in OOD detection, verifying its effectiveness in both preventing ID accuracy degradation and enhancing OOD detection performance in the context of image classification tasks.


\noindent\textbf{Diffusion models in OOD detection.}\quad Inspired by non-equilibrium thermodynamics, the Diffusion Probabilistic Model (DPM) was proposed to model the data distribution that enables precise sampling and evaluation of probabilities \cite{sohl2015deep}. Afterward, plenty of variants of the DPM have been proposed, such as DDPM \cite{ho2020denoising}, NCSNs \cite{NEURIPS2019_3001ef25}, DDPM++ \cite{song2021scorebased}, and so forth. Nowadays, DPM has achieved great successes in many sorts of vision tasks, including image generation \cite{saharia2022photorealistic, dockhorn2022scorebased}, image inpainting \cite{cao2023zits++, lugmayr2022repaint} and image segmentation \cite{wolleb2022diffusion, chen2023generative}, etc. In respect of OOD detection, many researchers propose to use diffusion models to perform outlier detection. For instance, in \cite{graham2023denoising}, the authors apply DDPMs to perform unsupervised OOD detection, which leverages the reconstruction error metrics to confirm whether an image is OOD. DIFFGUARD \cite{graham2023denoising} uses the guidance from semantic mismatch to better discern ID samples from the OOD ones. LMD \cite{liu2023unsupervised} utilizes the diffusion model's manifold mapping ability to conduct unsupervised OOD detection.

Within our research, we explore the application of Stable Diffusion \cite{rombach2022high} in synthesizing OOD data in the latent space.

\section{Preliminaries}
Let $\mathcal{X}$ denote the in-distribution input image space, $\mathcal{Y}=\left\{1, 2, \cdots, C\right\}$ denote the label space. We consider the multi-classification task using a standard dataset denoted by $D=\left\{(\textbf{x}_i, y_{i})\right\}_{i=1}^{N}$, which is drawn \textit{i.i.d.} from the joint data distribution $P_{\mathcal{X}\mathcal{Y}}^{in}$, where $N$ is the number of samples, $\textbf{x}_{i}\in\mathcal{X}, y_{i}\in\mathcal{Y}, i=1, \cdots, N$. 

\noindent\textbf{OOD detection.}\quad
When deploying a machine learning model in the real world, a reliable classifier should not only classify ID samples, but also accurately identify OOD samples. 

Formally, given a test sample $\mathbf{x}$ and a pre-trained classifier $f\left(\cdot\right)$, an OOD detection decision function can be written as 
\begin{equation*}
    \mathcal{G}\left(\mathbf{x}; f\right) = \left\{\begin{array}{cc}
      \text{ID},   & S\left(\mathbf{x}; f\right) \geq \alpha,\\
      \text{OOD},   & S\left(\mathbf{x}; f\right) < \alpha,
    \end{array}\right.
\end{equation*}
where $\alpha$ is the threshold, ID represents that sample $\mathbf{x}$ comes from the in-distribution dataset and OOD means that the sample $\mathbf{x}$ comes from the out-of-distribution dataset. $S(\mathbf{x};f)$ is the score function.



\noindent\textbf{Stable Diffusion model.}\quad 
To make the computation and training of diffusion models more efficient, in \cite{rombach2022high}, the authors proposed using an encoder $\mathcal{E}\left(\cdot\right)$ to compress the input images into a latent representation $\mathbf{z}=\mathcal{E}(\mathbf{x}), \mathbf{z}\in\mathbb{R}^{c\times h \times w}$. Then, they use a decoder $\mathcal{D}\left(\cdot\right)$ to decode the latent representation $\mathbf{z}$ to the image space, which can be formally written as $\hat{\mathbf{x}}=\mathcal{D}(\mathbf{z})$. In practical use, we synthesize images by using the text prompts' embeddings of label $y\in\mathcal{Y}$ for guidance.


\section{Methodology}

Our OAL training framework is shown in Figure \ref{OAL-pipeline}. In brief, it primarily addresses the following issues: (1) how to effectively collect and train with outliers synthesized by Stable Diffusion in the latent space; (2) how to mitigate the degradation of ID performance when training with the generated OOD data. 

The whole workflow of OAL consists of 2 stages. In Stage 1, we utilize the ID embeddings extracted from a pre-trained teacher model as input for the $k$-NN algorithm to sample class-conditioned OOD embeddings. Then, we use Stable Diffusion \cite{rombach2022high} to further generate OOD latent embeddings for OE training. In Stage 2, we develop MICL and IDKD modules to effectively regularize the model using the synthesized OOD data. Finally, we use the model trained by OAL to perform OOD inference. In the following sections, we will introduce our framework with equations and reveal more details of our contributions.

\subsection{Latent Space Outlier Synthesis}\label{sec:oe}
Following NPOS, we employ non-parametric nearest neighbor distance to extract OOD embeddings from a large vision model, without requiring distributional assumptions on ID feature. For the diffusion-based outlier synthesis process using Stable Diffusion, we utilize only its latent representation rather than the image.

\noindent\textbf{OOD features sampling using $k$-NN.}\quad
To reduce overall training time, we directly employ $k$-NN to sample outliers from the in-distribution features extracted by a pretrained vision transformer \cite{dosovitskiy2020image}, which has already achieved superior performance on ID image classification tasks.

Specifically, denote $\mathbb{Z}=\left\{z_1, z_2, \cdots, z_n\right\}$ as the set of the normalized in-distribution training data feature vectors extracted by the pre-trained transformers, $z_i\in \mathbb{R}^{D}$, $D$ is the dimension of the transformer's penultimate layer's feature space. Then, for any $z^{\prime}\in \mathbb{Z}$, we calculate the $k$-NN distance with respect to $\mathbb{Z}$, that is,
\begin{equation*}
    d_k\left( z^{\prime}, z_{k}\right) = \vert\vert z^{\prime} - z_k\vert\vert_2,
\end{equation*}
where $z_k$ is the $k$-th nearest embedding vectors in $\mathbb{Z}$, and $z_{k}^*$ is the vector with max $k$-NN distance. We select top-$k$ points with the largest $k$-NN distance as the boundary point of ID data. We denote these boundary points as $\Tilde{\mathbb{Z}}=\left\{z_{i}^*\right\}_{i=1}^k$. Afterwards, for each class $c\in \mathcal{Y}$, we take $\hat{z}_i^*:= \vert\vert\mathcal{T}(c)\vert\vert_F\cdot z_{i}^* ~(i=1, \cdots, k)$ as the center point of Gaussian kernels $\mathcal{N}(\hat{z}_i^*,\sigma^2\mathbf{I})$ to repeatedly sample class-conditioned OOD features from them, wherein $\mathcal{T}(c)$ is the text embedding of the name of class $c$, $\vert\vert\cdot\vert\vert_F$ denotes the Frobenius norm. We denote these OOD features sampled around $\hat{z}_i^*$ as $\mathbb{V}_i=\left\{\mathbf{v}_1, \cdots, \mathbf{v}_m\right\}~(i=1, \cdots, k)$. To ensure the collected points are far away from the ID data, we apply $k$-NN distance search again to select data in $\mathbb{V}_i$ with top-$k$ $k$-NN distance concerning $\mathbb{Z}$. 

We denote the final outliers sampled by $k$-NN as $\Tilde{\mathbb{V}}=\left\{\Tilde{\mathbf{v}}_1, \cdots, \Tilde{\mathbf{v}}_l\right\}$, $l$ is a positive integer.

\noindent\textbf{OOD latent embedding synthesis.}\quad
To reduce the computational cost of the Diffusion-based outlier synthesis approaches and obtain more diverse OOD samples for higher OOD detection performance \cite{mirzaeirodeo}, we further develop a latent-based outlier synthesis module using Stable Diffusion.

To be specific, we first take outlier embeddings $\Tilde{\mathbf{v}}\in \Tilde{\mathbb{V}}$ as the guidance to synthesize OOD images. However, in this study, we do not collect the final image outputs. Instead, we collect the latent representations $\mathbf{z}^{ood}$ that are output during the generation process of the OOD image, which can be formulated as follows:
\begin{equation*}
    \mathbf{z}^{ood}\sim \Tilde{P}\left(\mathbf{x}\vert \Tilde{\mathbf{v}}\right),
\end{equation*}
where $\mathbf{z}^{ood}\in \mathbb{R}^{c\times h\times w}$, $c$ is the number of channels, $h$ is the height of the generated latent representations, $w$ is the width of the generated representations, $\mathbf{x}$ is the input image, $\Tilde{P}\left(\mathbf{x}\vert \Tilde{\mathbf{v}}\right)$ is the distribution of the $\Tilde{\mathbf{v}}$ conditioned latent space.

After the generation step, we average the first two dimensions of $\mathbf{z}^{ood}$ to obtain the final OOD latent embeddings 
\begin{equation}\label{eq-mean}
    f^{ood} = \text{Mean}(\mathbf{z}^{ood}),
\end{equation}
to train the model. In equation (\ref{eq-mean}), $\text{Mean}(\cdot)$ denotes the operator that takes the average of the first two dimensions of $\mathbf{z}^{ood}$.

\subsection{OOD Aware Model Regularization}
To effectively regularize the model's decision boundary between ID and OOD features, we develop a contrastive learning module, named as MICL, instead of using energy-based uncertainty loss function \cite{du2022towards, tao2023nonparametric}. Moreover, we develop a knowledge distillation module, named IDKD, for boosting the model's in-distribution classification performance.

\begin{table*}[t]
    \centering
    \caption{Evaluation on CIFAR-100 benchmark. 
    We report standard deviations across 3 runs for our method. The bold numbers represent the best results. Here we find that OAL achieves the best result across various challenging OOD datasets.
    }
    \label{tab-cifar100}
    \resizebox{\textwidth}{!}{
    \begin{tabular}{cccccccccccccccc}
    \toprule[2.5pt]
        \multirow{3}{*}{\textbf{Methods}} &  \multicolumn{4}{c}{\textbf{Near-OOD}}  & \multicolumn{8}{c}{\textbf{Far-OOD}} & \multicolumn{2}{c}{} & \multirow{4}{*}{\textbf{ID ACC}} \\ 
        ~ & \multicolumn{2}{c}{\textbf{CIFAR-10}} & \multicolumn{2}{c}{\textbf{TIN}}  & \multicolumn{2}{c}{\textbf{MNIST}} & \multicolumn{2}{c}{\textbf{SVHN}} & \multicolumn{2}{c}{\textbf{Texture}} & \multicolumn{2}{c}{\textbf{Places365}} &\multicolumn{2}{c}{\textbf{Average}} & ~ \\
          ~ & \textbf{FPR@95 $\downarrow$} & \textbf{AUROC $\uparrow$} & \textbf{FPR@95 $\downarrow$} & \textbf{AUROC $\uparrow$} & \textbf{FPR@95 $\downarrow$} & \textbf{AUROC $\uparrow$} & \textbf{FPR@95 $\downarrow$} & \textbf{AUROC $\uparrow$} & \textbf{FPR@95 $\downarrow$} & \textbf{AUROC $\uparrow$} & \textbf{FPR@95 $\downarrow$} & \textbf{AUROC $\uparrow$} & \textbf{FPR@95 $\downarrow$} & \textbf{AUROC $\uparrow$} & ~\\
        \midrule[1.5pt]
        \multicolumn{16}{c}{\textit{Post hoc-based}} \\ 
        \midrule[1.5pt]
        MSP \cite{hendrycks2017a} & \textbf{58.66} & 78.88 & 62.74 & 79.26 & 48.37 & 80.25 & 46.70 & 84.48 & 80.96 & 68.70 & 49.90 & 83.57 & 57.89 & 79.19 & 76.22 \\  
        MDS \cite{lee2018simple} & 77.20 & 62.39 &  80.12 & 59.68 & 70.54 & 63.66 & 77.51 & 58.45 & 66.02 & 79.28 & 73.03 & 62.96 & 74.04 & 64.40 & 75.93\\ 
        Gram \cite{sastry2020detecting} & 91.87 & 51.72 & 100.00 & 18.37 &  94.62 & 40.23 & \textbf{12.32} & \textbf{97.34} & 100.00 & 38.73 & 100.00 & 11.86 & 83.14 & 43.04 & 76.22 \\   
        EBO \cite{liu2020energy} & 60.84 & 78.43 & 58.47 & 82.51 & 46.43 & 81.36 & 38.50 & 87.53 & 86.09 & 68.86 & 44.62 & 86.39 & 55.83  & 80.85 & 76.22 \\  
        ReAct \cite{sun2021react} & 71.14 & 73.38 & 58.07 & 83.74 &  58.37 & 77.55 & 34.61 & 89.12 & 76.80 & 70.22 & 47.08 & 86.80 & 57.68 & 80.14 & 75.36 \\   
        ViM \cite{wang2022vim} & 65.48 & 71.01 & 65.46 & 68.93 & 55.78 & 74.39 & 62.73 & 69.02 & \textbf{54.17} & \textbf{86.04} & 62.14 & 71.23 & 60.96 & 73.44 & 76.22 \\ 
        KNN \cite{sun2022out} & 72.37 & 76.10 & 55.08 & 83.40 & 45.54 & 84.91 & 50.30 & 85.71 & 46.22 & 85.65 & 44.61 & 86.27 & 52.35 & \textbf{83.67} & 76.22 \\  
        GEN \cite{liu2023gen} & 60.82 & 78.44 & 58.58 & 82.49 & 46.44 & 81.36 & 38.50 & 87.52 & 86.10 & 68.86 & 44.60 & 86.38 & 55.84 & 80.84 & 76.22 \\ 
        \midrule[1.5pt]
        \multicolumn{16}{c}{\textit{Outlier Synthesis-based}} \\ 
        \midrule[1.5pt]



        VOS \cite{du2022towards} & 64.59 & 76.94 & 59.37  & 81.36 & 55.36 & 79.91 & 47.51 & 84.82 & 88.32  & 63.76  &  46.66  & 85.04 &  60.30  & 78.64  & 74.04 \\  
        NPOS \cite{tao2023nonparametric}  & 97.96 & 46.23 & 93.78 & 54.13 & 85.53 & 49.58 & 89.04 & 57.00 & 90.26 & 60.12 & 94.30 & 46.09 & 91.81 & 52.19 & -  \\  
         OAL (Ours) & 61.24$\pm$2.85 & \textbf{79.33$\pm$0.58} & \textbf{54.25$\pm$8.39}   & \textbf{84.72$\pm$2.12}  &  \textbf{38.15$\pm$2.60} & \textbf{86.82$\pm$0.95} & 42.29$\pm$7.94 & 86.13$\pm$3.47 & 70.73$\pm$7.64 & 75.57$\pm$2.57 &  \textbf{41.14$\pm$3.20}  & \textbf{87.62$\pm$1.04} & \textbf{51.30$\pm$4.14}  & 83.36$\pm$1.43 & \textbf{76.89$\pm$0.20}\\  

        \bottomrule[2.5pt]
    \end{tabular}
    }
 
\end{table*}

\noindent\textbf{Mutual information-based contrastive learning.}\quad
Inspired by CLUB \cite{cheng2020club}, we develop a mutual information-based contrastive learning module to enlarge the discrepancy between ID and OOD data in the feature space. To be specific, we propose a learning technique that minimizes the upper bound of mutual information between ID and OOD features, making them more independent of each other. We name it as MICL (Mutual Information Contrastive Learning). In our work, we apply the theory proposed by CLUB to regularize the model using the synthesized OOD data.

To this end, we use mutual information to quantify the distance between the ID and OOD feature distribution, which takes the form as follows,
\begin{align*}
    \Tilde{\mathbf{u}} &= \phi_{\theta}(\Tilde{\mathbf{v}}), \\
    I(\Tilde{\mathbf{u}}; f^{in})&=\mathbb{E}_{p(\Tilde{\mathbf{u}}, f^{in})}\left[\log\frac{p(\Tilde{\mathbf{u}}, f^{in})}{p(\Tilde{\mathbf{u}})p(f^{in})}\right],
\end{align*}
where $f^{in}\in \mathbb{R}^{d}$ is the penultimate layer's feature of the student network, which is the ResNet-18 shown in Figure \ref{OAL-pipeline}, $\Tilde{\mathbf{v}}\in \Tilde{\mathbb{V}}$ denotes the OOD embeddings sampled by $k$-NN, $\phi_{\theta}:\mathbb{R}^D\rightarrow\mathbb{R}^d$ is a fully connected neural network that aligns $\Tilde{\mathbf{v}}$ and $f^{in}$ into the same feature space. Similarly, the mutual information between outliers $f^{ood}\in\mathbb{R}^{w}$ synthesized by Stable Diffusion and $f^{in}$ can be written as the following,
\begin{align*}
    \Tilde{f}^{ood} &= \phi_{\theta}^\prime(f^{ood}), \\
    I(\Tilde{f}^{ood}; f^{in})&=\mathbb{E}_{p(\Tilde{f}^{ood}, f^{in})}\left[\log\frac{p(\Tilde{f}^{ood}, f^{in})}{p(\Tilde{f}^{ood})p(f^{in})}\right],
\end{align*}
where $\phi^\prime_{\theta}:\mathbb{R}^w\rightarrow\mathbb{R}^d$ is also a fully connected neural network that projects $f^{ood}$ into the same feature space as $f^{in}$. 

Let 
\begin{align}
\mathcal{L}_{\text{MI}}(f^{in}, \Tilde{\mathbf{u}})&:=\hat{I}(\Tilde{\mathbf{u}}; f^{in}) ~~~~~~~~~~~~~~~~~~~~~~~~~~~~~~~~~~ (\text{MICL1}) \notag\\
    &= \mathbb{E}_{p(\Tilde{\mathbf{u}},f^{in})}[\log p{(f^{in}|\Tilde{\mathbf{u}})}] \notag\\
    &-\mathbb{E}_{p(\Tilde{\mathbf{u}})}\mathbb{E}_{p(f^{in})}[\log p(f^{in}|\Tilde{\mathbf{u}})], \label{mi-loss1} \\
\mathcal{L}_{\text{MI}}(f^{in}, \Tilde{f}^{ood})&:=\hat{I}(\Tilde{f}^{ood}; f^{in}) ~~~~~~~~~~~~~~~~~~~~~~~~~~~~~ (\text{MICL2})\notag\\
    &= \mathbb{E}_{p(\Tilde{f}^{ood},f^{in})}[\log p(f^{in}|\Tilde{f}^{ood})] \notag \\
    &-\mathbb{E}_{p(\Tilde{f}^{ood})}\mathbb{E}_{p(f^{in})}[\log p(f^{in}|\Tilde{f}^{ood})]. \label{mi-loss2} 
\end{align}

We use MICL1 and MICL2, the mutual information estimators, as the loss functions to regularize the model with synthesized data. 

\begin{table*}[t]

    \caption{Comparisons on CIFAR-10 benchmark. 
    We report standard deviations across 3 runs for our method. The bold numbers represent the best results. In this benchmark, OAL surpasses many other algorithms on average FPR@95 and AUROC with a certain margin.}
    \label{tab-cifar10}
    \resizebox{\textwidth}{!}{
    \begin{tabular}{cccccccccccccccc}
    \toprule[2.5pt]
        \multirow{3}{*}{\textbf{Methods}} &  \multicolumn{4}{c}{\textbf{Near-OOD}}  & \multicolumn{8}{c}{\textbf{Far-OOD}} & \multicolumn{2}{c}{} & \multirow{4}{*}{\textbf{ID ACC}} \\ 
        ~ & \multicolumn{2}{c}{\textbf{CIFAR-100}} & \multicolumn{2}{c}{\textbf{TIN}}  & \multicolumn{2}{c}{\textbf{MNIST}} & \multicolumn{2}{c}{\textbf{SVHN}} & \multicolumn{2}{c}{\textbf{Texture}} & \multicolumn{2}{c}{\textbf{Places365}} &\multicolumn{2}{c}{\textbf{Average}} & ~ \\
          ~ & \textbf{FPR@95 $\downarrow$} & \textbf{AUROC $\uparrow$} & \textbf{FPR@95 $\downarrow$} & \textbf{AUROC $\uparrow$} & \textbf{FPR@95 $\downarrow$} & \textbf{AUROC $\uparrow$} & \textbf{FPR@95 $\downarrow$} & \textbf{AUROC $\uparrow$} & \textbf{FPR@95 $\downarrow$} & \textbf{AUROC $\uparrow$} & \textbf{FPR@95 $\downarrow$} & \textbf{AUROC $\uparrow$} & \textbf{FPR@95 $\downarrow$} & \textbf{AUROC $\uparrow$} & ~\\
        \midrule[1.5pt]
        \multicolumn{16}{c}{\textit{Post hoc-based}} \\ 
        \midrule[1.5pt]
        MSP \cite{hendrycks2017a} & 34.81 & 89.52 & 25.43 & 92.18 & 22.74 & 92.94 & 16.66 & 94.49 & 31.97 & 89.46 & 24.24 & 92.69 & 25.98 & 91.88 & 94.66 \\  
        MDS \cite{lee2018simple} & 75.80 & 67.70 & 61.78 & 77.06 & 39.36 & 85.83 & 72.61 & 64.22 & 25.97 & 95.06 &  62.92 & 76.32 & 56.41 & 77.70 & 94.68  \\ %
        Gram \cite{sastry2020detecting} & 86.06 & 65.03 &  99.99 &  23.78 & 86.56 & 44.86 & 10.52 & \textbf{97.71} &  100.00  & 32.81  & 100.00 & 33.20 & 80.52 & 49.57 & 94.66 \\   
        EBO \cite{liu2020energy} & 34.40 & 90.65 &  21.83 & 94.25 & 19.68 & 94.74 & 12.24 & 96.62 & 30.87 & 90.91 &  20.28  & 94.88 & 23.22 & 93.68 & 94.66 \\  
        ReAct \cite{sun2021react} & 40.78 & 89.67 &  25.48  & 93.54  & 23.13 & 94.06 & 14.50 & 96.08 & 35.63 & 89.96 &  23.52 & 94.22 & 27.17 & 92.92 & 94.64 \\   
        ViM \cite{wang2022vim} & 36.97 & 88.71 & 21.32 & 94.16 & 17.50 & 95.70 & 21.42 & 91.34 &  \textbf{9.88} & \textbf{98.24} & 21.87 & 94.09 & 21.49 & 93.71 & 94.66 \\ 
        KNN \cite{sun2022out} & 37.37 & 89.45 & 30.39  & 92.03 & 23.28 & 93.45 & 26.39 & 91.54 & 32.80 & 90.10 & 27.41 & 92.96 & 29.61 & 91.59 & 94.66 \\  
        GEN \cite{liu2023gen} & 34.38 & 90.65 &  22.39 & 94.16 & 19.98 & 94.68 & 12.61 & 96.55 & 31.10 & 90.88 & 20.50 & 94.78 &  23.49 & 93.62 & 94.66 \\ 
        \midrule[1.5pt]
        \multicolumn{16}{c}{\textit{Outlier Synthesis-based}} \\ 
        \midrule[1.5pt]

        VOS \cite{du2022towards} & 34.83 & 90.51 & 21.40 & 94.19 & 13.79 & 96.44 & 13.79 & 96.18 &  29.76  & 90.73 &  19.72 & 94.77 & 22.22 & 93.80 & 94.28 \\  
        NPOS \cite{tao2023nonparametric}  & 92.57 & 60.19 & 90.89 & 49.45 & 86.47 & 56.38 & 37.18 & 90.29 & 79.24 & 63.85 & 97.62 & 44.92 & 80.66 & 60.85 &  - \\ 
        OAL (Ours) & \textbf{31.22$\pm$0.96} & \textbf{91.48$\pm$0.31} & \textbf{21.07$\pm$0.62} & \textbf{94.49$\pm$0.08} & \textbf{5.67$\pm$0.29} & \textbf{98.52$\pm$0.08} & \textbf{9.62$\pm$0.40} & 97.30$\pm$0.13 & 21.26$\pm$0.49  & 93.82$\pm$0.23 &  \textbf{18.52$\pm$0.24} & \textbf{95.49$\pm$0.02} & \textbf{17.89$\pm$0.01} & \textbf{95.19$\pm$0.03} & \textbf{94.71$\pm$0.18} \\  
        \bottomrule[2.5pt]
    \end{tabular}
    }
    
\end{table*}

Indeed, one can directly show that $\hat{I}(\mathbf{u};f^{in})$ and $\hat{I}(\Tilde{f}^{ood};f^{in})$ are upper bounds of $I(\mathbf{u};f^{in})$ and $I(\Tilde{f}^{ood};f^{in})$ by calculating the gaps between them. We summarize this fact in the following Theorem.

\begin{theorem}
\label{thm-4}
Based on the discussions above, we have the following inequalities,
\begin{itemize}
    \item (\romannumeral 1) $I(\Tilde{\mathbf{u}}; f^{in}) \leqslant \hat{I}(\Tilde{\mathbf{u}}; f^{in})$ with equality if and only if $\Tilde{\mathbf{u}}$ and $f^{in}$ are independent;
    \item (\romannumeral 2) $I(\Tilde{f}^{ood}; f^{in}) \leqslant \hat{I}(\Tilde{f}^{ood}; f^{in})$ with equality if and only if 
    $\Tilde{f}^{ood}$ and $f^{in}$ are independent.
\end{itemize}

\end{theorem}

\begin{proof}
For the first inequality in Theorem 1, $I$ and $\hat{I}$ can be written as the following form:
\begin{align*}
I(\Tilde{\mathbf{u}}; f^{in})&=\mathbb{E}_{p(\Tilde{\mathbf{u}}, f^{in})}\left[\log\frac{p(\Tilde{\mathbf{u}}, f^{in})}{p(\Tilde{\mathbf{u}})p(f^{in})}\right],\\
\hat{I}(\Tilde{\mathbf{u}}; f^{in})&=\mathbb{E}_{p(\Tilde{\mathbf{u}},f^{in})}[\log p{(f^{in}|\Tilde{\mathbf{u}})}] \\
    &-\mathbb{E}_{p(\Tilde{\mathbf{u}})}\mathbb{E}_{p(f^{in})}[\log p(f^{in}|\Tilde{\mathbf{u}})].
\end{align*}

Denote 
\begin{equation*}
   \Delta:=\hat{I}(\Tilde{\mathbf{u}}; f^{in}) - I(\Tilde{\mathbf{u}}; f^{in}),
\end{equation*}
then, we have

\begin{align*}
 \Delta&=\mathbb{E}_{p(\Tilde{\mathbf{u}},f^{in})}[\log p{(f^{in}|\Tilde{\mathbf{u}})}] -\mathbb{E}_{p(\Tilde{\mathbf{u}})}\mathbb{E}_{p(f^{in})}[\log p(f^{in}|\Tilde{\mathbf{u}})]\\
& -\mathbb{E}_{p(\Tilde{\mathbf{u}}, f^{in})}\left[\log\frac{p(\Tilde{\mathbf{u}}, f^{in})}{p(\Tilde{\mathbf{u}})p(f^{in})}\right] \\
& =\left[\mathbb{E}_{p(\Tilde{\mathbf{u}},f^{in})}[\log p{(f^{in}|\Tilde{\mathbf{u}})}] -\mathbb{E}_{p(\Tilde{\mathbf{u}})}\mathbb{E}_{p(f^{in})}[\log p(f^{in}|\Tilde{\mathbf{u}})]\right] \\
&-\left[\mathbb{E}_{p(\Tilde{\mathbf{u}},f^{in})}[\log p(f^{in}|\Tilde{\mathbf{u}})]-\mathbb{E}_{p(\Tilde{\mathbf{u}},f^{in})}[\log p(f^{in})]\right] \\
&=\left[\mathbb{E}_{p(f^{in})}[\log p(f^{in})]-\mathbb{E}_{p(\Tilde{\mathbf{u}})}\mathbb{E}_{p(f^{in})}[\log p(f^{in}|\Tilde{\mathbf{u}})]\right] \\
&- \left[\mathbb{E}_{p(\Tilde{\mathbf{u}},f^{in})}[\log p(f^{in}|\Tilde{\mathbf{u}}]-\mathbb{E}_{p(\Tilde{\mathbf{u}},f^{in})}[\log p(f^{in}|\Tilde{\mathbf{u}}]\right]\\
&=\mathbb{E}_{p(f^{in})}[\log p(f^{in})]-\mathbb{E}_{p(\Tilde{\mathbf{u}})}\mathbb{E}_{p(f^{in})}[\log p(f^{in}|\Tilde{\mathbf{u}})] \\
&= \mathbb{E}_{p(\Tilde{\mathbf{u}})p(f^{in})}\left[\log\frac{p(f^{in})}{p(f^{in}\vert \Tilde{\mathbf{u}})}\right]\\
&=\mathbb{E}_{p(\Tilde{\mathbf{u}})p(f^{in})}\left[\log\frac{p(\Tilde{\mathbf{u}})p(f^{in})}{p(\Tilde{\mathbf{u}}, f^{in})}\right]\\
&=\text{KL}\left(p(\Tilde{\mathbf{u}})p(f^{in})\vert\vert p(\Tilde{\mathbf{u}}, f^{in})\right).
\end{align*}

Herein, $\text{KL}(\cdot\vert\vert\cdot)$ denotes the  Kullback–Leibler (KL) divergence. By Theorem 6.2.1 in \cite{pml1Book}, we have $\Delta\geq 0$. Therefore, $\hat{I}(\Tilde{\mathbf{u}}; f^{in})$ is an upper bound of $I\left(\Tilde{\mathbf{u}}; f^{in}\right)$. Meanwhile, the equality achieves if and only if $p(\Tilde{\mathbf{u}})p(f^{in})=p(\Tilde{\mathbf{u}}, f^{in})$, i.e., $\Tilde{\mathbf{u}}$ and $f^{in}$ are independent.

Similar calculation can be done for the second inequality in Theorem 1. We omit it here.
\end{proof}







\noindent\textbf{In-distribution performance boosting by knowledge distillation.}\quad \label{sec:kd}
Besides using mutual information-based loss function to enlarge feature discrepancies, we develop \textbf{I}n-\textbf{D}istribution \textbf{K}nowledge \textbf{D}istillation module to prevent the model's ID accuracy degradation when training with outliers while preserving the model's inference speed \cite{vengertsev2023confidence}. We refer to this module as IDKD for brevity.

In IDKD, we first leverage KL divergence loss to allow the student network to mimic the teacher network's logits outputs \cite{hinton2015distilling}, which can be formulated as follows:
\begin{align*}
    \mathcal{L}_{\text{KL}}^{1}\left(\hat{y}_{S}^\prime, \hat{y}_{T}\right)&=\int_{\mathbf{x}\in\mathcal{X}}\ln\left(\frac{\hat{y}_{S}^\prime}{\hat{y}_{T}}\right) \hat{y}_{S}^\prime d\mathbf{x}, \\
    \hat{y}_S^\prime &= \text{Softmax}(\hat{y}_S),
\end{align*}
where $\hat{y}_{S}$ is the logits output of the student model, $\hat{y}_{T}$ is the softmax prediction of the teacher model.

Then, to transfer ID information at the feature level \cite{lin2022knowledge,chen2022knowledge}, we develop a domain transfer network consisting of a stack of Multi-Layer Perceptrons (MLPs) to align the teacher model's feature spaces with the student model's penultimate layer's feature space. Formally, we denote the domain transfer network as $\varphi\left(\cdot\right):\mathbb{R}^D\rightarrow \mathbb{R}^d, D>d$. Then, the learning objective function for minimizing the distribution discrepancy between the teacher model's feature and the student model's feature can be written as follows:
\begin{align*}
    \mathcal{L}_{\text{KL}}^{2}\left(\Tilde{f}^{in}, \Tilde{f}_T^{in}\right)&=\int_{\mathbf{x}\in\mathcal{X}}\ln\left(\frac{\Tilde{f}^{in}}{\Tilde{f}_T^{in}}\right) \Tilde{f}^{in} d\mathbf{x}, \\
    \Tilde{f}_T^{in} &= \text{Softmax}(\varphi(f^{in}_T)), \\
    \Tilde{f}^{in} &= \text{Softmax}(f^{in}),
\end{align*}
where $f^{in}$ is the in-distribution feature of the student model, $f_T^{in}$ is the ID feature extracted by the teacher model.

Finally, the total loss function takes the form as follows:
\begin{align}
    \mathcal{L}_{total} &= \mathcal{L}_{\text{CE}}\left(\hat{y}_S, y_{gt}\right)+\alpha_1\mathcal{L}_{\text{KL}}^{1}\left(\hat{y}_{S}^\prime, \hat{y}_{T}\right)\notag\\
    &+ \alpha_2\mathcal{L}_{\text{KL}}^{2}\left(\Tilde{f}^{in}, \Tilde{f}_T^{in}\right)+\beta\mathcal{L}_{\text{MI}}(f^{in}, \Tilde{\mathbf{u}})\notag\\
    &+\gamma\mathcal{L}_{\text{MI}}(f^{in}, \Tilde{f}^{ood}), \label{loss-func}
\end{align}
where $y_{gt}$ is the ground truth, $\alpha_1, \alpha_2, \beta, \gamma$ are the weights of each component of $\mathcal{L}_{total}$, $\mathcal{L}_{\text{CE}}$ is the cross-entropy loss, $\mathcal{L}_{\text{KL}}^{1}$ is the KL-divergence loss between the logits output of the student and teacher network, $\mathcal{L}_{\text{KL}}^{2}$ is the KL-divergence loss between the feature output of student and teacher network.

\begin{table*}[t]
    \centering
    \caption{Comparisons with other outlier synthesis-based methods using CIFAR-100 as the ID data. $\dagger$ represents that the method is reproduced by ourselves using the official code released by the authors. 
    }
    \label{tab-sota-1}

\resizebox{0.68\textwidth}{!}{
    \begin{tabular}{cccccccc}
    \toprule[2.5pt]
        \multirow{2}{*}{\textbf{Methods}}  &\multicolumn{2}{c}{\textbf{SVHN}} 
        &\multicolumn{2}{c}{\textbf{Texture}}
        & \multicolumn{2}{c}{\textbf{Places365}} 
& \multirow{2}{*}{\textbf{ID ACC}}
        
        \\
          ~ & \textbf{FPR@95 $\downarrow$} & \textbf{AUROC $\uparrow$}& \textbf{FPR@95 $\downarrow$} & \textbf{AUROC $\uparrow$}& \textbf{FPR@95 $\downarrow$} & \textbf{AUROC $\uparrow$}&~
          \\
        \midrule[1.5pt]
    $\text{GAN}$~ \cite{lee2018training} 
    & 89.45 & 66.95 & 92.80 & 62.99 & 88.75 & 66.76 & 70.12
    \\  
    $\text{ATOL}^{\dagger}$~ \cite{NEURIPS2023_e43f900f} 
    & 77.20 & 79.05 & 62.80 & 83.95 & 75.65 & 78.79 & 72.99
    \\
    $\text{Dream-OOD}$~ \cite{NEURIPS2023_bf5311df} 
    & 58.75$\pm$0.6 & \textbf{87.01$\pm$0.1}  & \textbf{46.60 $\pm$ 0.4} & \textbf{88.82 $\pm$ 0.7} & 70.85$\pm$1.6 & 79.94$\pm$0.2 & \textbf{78.94}
    \\ 
    \midrule[1.5pt]
     OAL (Ours) 
     & \textbf{42.29$\pm$7.94} & 86.13$\pm$3.47 & 70.73$\pm$7.64 & 75.57$\pm$2.57 &  \textbf{41.14$\pm$3.20}  & \textbf{87.62$\pm$1.04} & 76.89$\pm$0.20\\  

        \bottomrule[2.5pt]
    \end{tabular}
}
\end{table*}








\section{Experiments}
\subsection{Experimental Setup}\label{data-setup}
\noindent\textbf{Datasets.}\quad 
We evaluate our methods on two commonly used benchmarks: CIFAR-10 and CIFAR-100 \cite{krizhevsky2009learning}. On the CIFAR-10 benchmark, we take CIFAR-10 \cite{krizhevsky2009learning} as the ID data, CIFAR-100 \cite{krizhevsky2009learning}, TIN (Tiny ImageNet) \cite{le2015tiny}, MNIST \cite{deng2012mnist}, SVHN \cite{netzer2011reading}, Texture \cite{cimpoi2014describing}, Places365 \cite{zhou2017places} as the OOD datasets. When using CIFAR-100 \cite{krizhevsky2009learning} as the ID dataset, we use CIFAR-10 \cite{krizhevsky2009learning}, TIN \cite{le2015tiny}, MNIST \cite{deng2012mnist}, SVHN \cite{netzer2011reading}, Texture \cite{cimpoi2014describing}, Places365 \cite{zhou2017places} as the OOD datasets. 

\noindent\textbf{Evaluation metrics.}\quad To evaluate the performance of OOD detection, we use 3 commonly used metrics following \cite{NEURIPS2023_bf5311df, tao2023nonparametric}: (\romannumeral 1) the False Positive Rate (FPR) at 95\% True Positive Rate; (\romannumeral 2) the Area Under the Receiver Operating Characteristic curve (AUROC); (\romannumeral 3) the in-distribution classification accuracy (ID ACC) of the tested models. 

\noindent\textbf{Implementation details.}\quad  We utilize ResNet-18 \cite{he2016deep} as the backbone of the student model, ViT-B/16 \cite{dosovitskiy2020image} as the backbone of the teacher model. We run all experiments using Python 3.8.19, Pytorch 1.13.1 with one V100 GPU. Our code is developed mainly based on OpenOOD \cite{yang2022openood,zhang2024openood}. To facilitate the implementation of knowledge distillation, we set both the train and test image resolution to $224\times 224$ for all methods on CIFAR-10/100 \cite{krizhevsky2009learning} benchmarks. To ensure fairness, the train batch size of all methods is set to 128 and the validation and test batch size are set to 200. For both CIFAR-10 and CIFAR-100 datasets \cite {krizhevsky2009learning}, the training epoch of our method is set to 100 and the learning rate is set to 0.005 using SGD \cite{robbins1951stochastic} as the optimizer. The weights in loss function (\ref{loss-func}) for both CIFAR-10 and CIFAR-100 are set to $\alpha_1=4.0$, $\alpha_2=8.0$, $\beta=0.1$, $\gamma=0.2$, respectively. We generate 300 OOD latent embeddings using Stable Diffusion v1.4 taking CIFAR-10/100 as the ID data. In addition, we do not apply any test time augmentation or data augmentation techniques.

\noindent\textbf{Baselines.}\quad  To evaluate the post hoc-based baselines, such as MSP \cite{hendrycks2017a}, ODIN \cite{liang2018enhancing}, Gram \cite{sastry2020detecting}, EBO \cite{liu2020energy}, ReAct \cite{sun2021react}, ViM \cite{wang2022vim}, KNN \cite{sun2022out}, GEN \cite{liu2023gen}, we directly use the settings given in OpenOOD \cite{yang2022openood,zhang2024openood} with the same pre-trained ResNet-18 backbone. In ODIN, the temperature is set to $T=1000$. For EBO, the temperature is set to $T=1$. For both VOS \cite{du2022towards} and NPOS \cite{tao2023nonparametric}, we sample 1000 points for each class following the original settings given in OpenOOD \cite{yang2022openood,zhang2024openood}. For both Dream-OOD \cite{NEURIPS2023_bf5311df} and GAN \cite{lee2018training}, we directly cite the results given in Dream-OOD \cite{NEURIPS2023_bf5311df}. Moreover, we reproduce ATOL \cite{NEURIPS2023_e43f900f} by using the official released code on our own device.


\subsection{Main Results}

\begin{table}[t] 
\centering
\caption{Ablation of the modules in OAL on the CIFAR-100 benchmark using EBO score.} 
\label{tab-ab2}
    \resizebox{0.43\textwidth}{!}{
\begin{tabular}{c|ccc|c|c|c}
  \toprule[2.5pt]
    & \makecell[c]{IDKD} & \makecell[c]{MICL1} & \makecell[c]{MICL2} & \textbf{FPR@95}$\downarrow$ & \textbf{AUROC} $\uparrow$ & \textbf{ID ACC}
  \\
  \midrule[1.5pt]
  (\romannumeral 1)& \XSolidBrush & \XSolidBrush & \XSolidBrush   & 55.83  & 80.85 & 76.22 \\
  (\romannumeral 2)&  \Checkmark & \XSolidBrush & \XSolidBrush & 51.45 & 83.50 & 77.42 \\
  (\romannumeral 3)&  \Checkmark  & \Checkmark & \XSolidBrush  & 48.72  & 84.41 & \textbf{77.66} \\
  (\romannumeral 4)& \XSolidBrush & \Checkmark & \Checkmark & 55.93 & 80.12 & 75.06 \\
  (\romannumeral 5-1)&  \Checkmark & \Checkmark & \Checkmark & \textbf{47.00}  & \textbf{84.91} & 76.76 \\
  (\romannumeral 5-2)&  \Checkmark & \Checkmark & \Checkmark & 51.30$\pm$4.14  & 83.36$\pm$1.43 & 76.89$\pm$0.20\\

  \bottomrule[2.5pt]
\end{tabular}
}
\end{table}


To demonstrate the efficacy of our methods, we apply OAL to the EBO \cite{liu2020energy} score and test the method on CIFAR-100 and CIFAR-10 \cite{krizhevsky2009learning} benchmarks. The other outlier synthesis-based methods, listed in Tables \ref{tab-cifar100} and \ref{tab-cifar10}, are \textit{all tested using ResNet-18 as the backbone and EBO as the score function}. 

\noindent\textbf{Evaluation on CIFAR-100 benchmark.}\quad
From Table \ref{tab-cifar100}, we can observe that OAL outperforms other outlier exposure-based methods, such as VOS \cite{du2022towards} and NPOS \cite{tao2023nonparametric}, as well as post hoc methods (e.g., MSP, MDS, ViM) with a certain margin. Specifically, we can observe that OAL outperforms VOS by 9.00 $\%$ in mean FPR@95 and by 4.72 $\%$ in mean AUROC. It also outperforms NPOS by 40.51$\%$ in mean FPR@95 and by 31.17$\%$ in mean AUROC. Besides, OAL surpasses EBO by 4.53$\%$ in FPR@95 and by 2.51 $\%$ in AUROC, averaged across the six OOD datasets.

\noindent\textbf{Evaluation on CIFAR-10 benchmark.}\quad
As shown in Table \ref{tab-cifar10}, OAL also achieves the best performance among all post hoc and outlier synthesis-based approaches. To be specific, OAL outperforms VOS by 4.33$\%$ in mean FPR@95 and by 1.39 $\%$ in mean AUROC. It also outperforms EBO by 5.33 $\%$ in mean FPR@95 and by 1.51 $\%$ in mean AUROC.

\begin{figure*}[t]
\centerline{\includegraphics[width=0.93\linewidth]{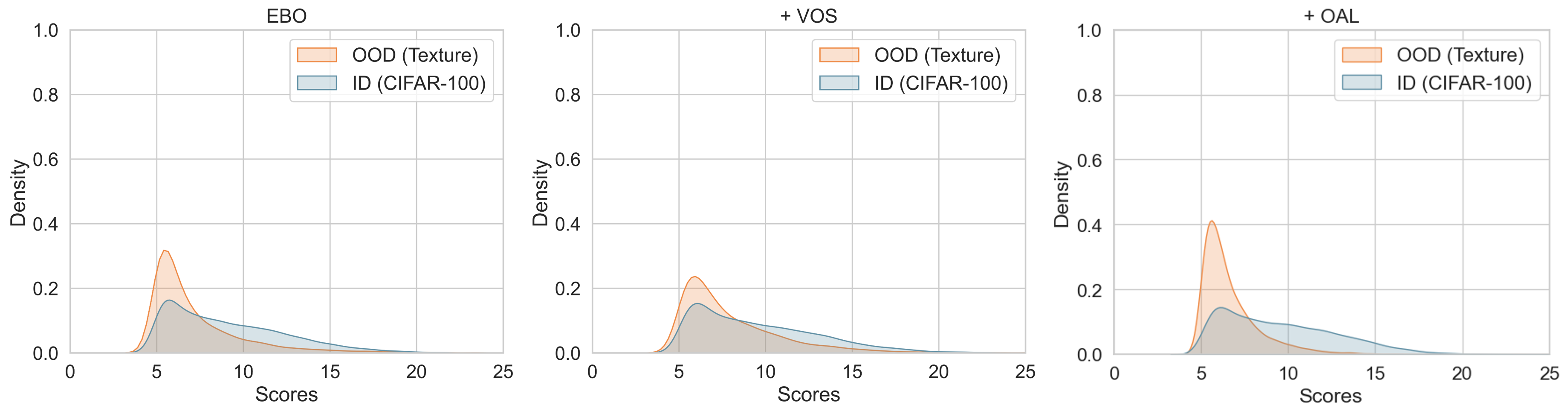}}
\caption{The score distribution visualization of EBO taking CIFAR-100 as the ID data, Texture as the OOD data and ResNet-18 as the backbone. As shown in the above figure, OAL has better OOD detection performance enhancement ability than VOS.
}
\Description{5}
\label{ebo-score-distribution}
\end{figure*}

\noindent\textbf{Comparisons with other outlier synthesis-based approaches.}\quad
Furthermore, we perform comparisons between OAL and other outlier synthesis-based methods, such as GAN \cite{lee2018training}, Dream-OOD \cite{NEURIPS2023_bf5311df}, and ATOL \cite{NEURIPS2023_e43f900f}, in Table \ref{tab-sota-1}. As shown in Table \ref{tab-sota-1}, OAL's mean FPR@95 is $16.49\%$ lower than Dream-OOD's on SVHN. Besides, OAL outperforms Dream-OOD by 29.71$\%$ in FPR@95 and by 7.68$\%$ in AUROC on Places365.

\noindent\textbf{Comparison of computational costs.}\quad Herein, we test the computational cost of OAL and other baselines on the CIFAR-100 benchmark. The results are summarized as follows: the post hoc methods require training a model on in-distribution data for 2.4 hours; VOS takes 12.6 hours for training; in Dream-OOD, as the authors stated in the paper, their approach requires learning the text-conditioned latent space for about 8.2 hours, 10.1 hours for generating 100K images, and 8.5 more hours to train with synthesized OOD data; while our approach only requires 0.5 hours to generate the outliers and the train time of our methods OAL takes about 8.7 hours. This result further supports our statement that synthesizing OOD data in the latent space using Stable Diffusion benefits the overall outlier exposure training process.

\begin{table}[t]
    \centering
    \caption{Ablation analysis of the backbone, taking CIFAR-100 as the in-distribution dataset.
    }
    \label{tab-backbone}
\resizebox{0.4\textwidth}{!}{
    \begin{tabular}{ccccc}
    \toprule[2.5pt]
        \multirow{2}{*}{\textbf{Backbone}} & \multirow{2}{*}{\textbf{Methods}}& \multicolumn{2}{c}{\textbf{Average}} & \multirow{2}{*}{\textbf{ID ACC}} \\ 
          ~ & ~ &\textbf{FPR@95 $\downarrow$} & \textbf{AUROC $\uparrow$} & ~\\
        \midrule[1.5pt]
        \multirow{2}{*}{ResNet-18} & EBO \cite{liu2020energy} & 55.83  & 80.85 & 76.22\\
        ~&  + OAL &\textbf{51.30$\pm$4.14}  & \textbf{83.36$\pm$1.43} & \textbf{76.89$\pm$0.20}\\
        \midrule[1.5pt]
        \multirow{2}{*}{ResNet-50} & EBO \cite{liu2020energy} & 59.55  &  77.31  &  75.16 \\
        ~&  + OAL & \textbf{40.08$\pm$2.87} & \textbf{88.74$\pm$1.33} & \textbf{86.37$\pm$0.34} \\
        \bottomrule[2.5pt] 
    \end{tabular}
}
\end{table}

\subsection{Ablation Study}\label{sec:ablation}  

\noindent\textbf{Ablation of the learning framework.}\quad
Here we investigate the impact of each component of our learning framework. The results are shown in Table \ref{tab-ab2}. We report average FPR@95 and AUROC on 6 OOD datasets in the CIFAR-100 \cite{krizhevsky2009learning} benchmark. In Table \ref{tab-ab2}, setting (\romannumeral 5-1) represents the best results of OAL across 3 training runs, and setting (\romannumeral 5-2) reports the mean $\pm$ standard deviation value of OAL across 3 training runs. 
As shown in Table \ref{tab-ab2},
the comparison between setting (\romannumeral 3) and setting (\romannumeral 5-1) shows the effectiveness of using OOD data synthesized by Diffusion for outlier exposure training. By comparing setting (\romannumeral 4) with setting (\romannumeral 5-1) and (\romannumeral 5-2), we can find that IDKD improves both the ID accuracy and the OOD detection performance of the model to a great extent. By comparing setting (\romannumeral 2) with (\romannumeral 5-1) and (\romannumeral 5-2), we can find that the combination of IDKD and MICL blocks significantly improves the OOD detection results.

\noindent\textbf{Ablation on regularization loss functions.}\quad
Here we perform ablation on the regularization loss of OAL. The results are presented in Table \ref{tab-rl-ablation}. As shown in Table \ref{tab-rl-ablation}, we can observe that using only MICL modules (the fourth row in the table) can obtain better OOD performance in FPR@95 and AUROC than OAL w/ energy loss, and VOS, which empirically demonstrated the regularization ability of our mutual information-based loss is better than the methods using energy-based loss functions. Additionally, by comparing the performance of OAL w/o IDKD and OAL, we can find that knowledge distillation approaches do have the ability to improve the model's OOD detection performance while maintaining its ID accuracy.

\begin{table}[t]
    \centering
    \caption{Ablation of the regularization loss on CIFAR-100 benchmark using ResNet-18 and EBO score.
    }
    \label{tab-rl-ablation}
    \resizebox{0.35\textwidth}{!}{
    \begin{tabular}{@{}cccc@{}}
    \toprule[2.5pt]
        \multirow{2}{*}{\textbf{Methods}} & \multicolumn{2}{c}{\textbf{Average}} & \multirow{2}{*}{\textbf{ID ACC}} \\ 
        ~& \textbf{FPR@95 $\downarrow$} & \textbf{AUROC $\uparrow$} &~ \\ 
        \midrule[1.5pt]
        EBO \cite{liu2020energy} & 55.83  & 80.85 & 76.22 \\
        VOS \cite{du2022towards} & 60.30  & 78.64  & 74.04 \\
        OAL w/ energy loss & 59.67 & 79.54 & 75.33 \\
        OAL w/o IDKD & 55.93 & 80.12 & 75.06\\
        OAL & \textbf{51.30$\pm$4.14}  & \textbf{83.36$\pm$1.43} & \textbf{76.89$\pm$0.20} \\  
        \bottomrule[2.5pt]
    \end{tabular}
    }
\end{table}

\noindent\textbf{Ablation on the backbone.}\quad
In this section, we analyze the impact of changes to the backbone. The results are shown in Table \ref{tab-backbone}. From Table \ref{tab-backbone} we can find that OAL is effective for both ResNet-18 and ResNet-50 using EBO score as the OOD detector, verifying that OAL is robust to backbone change.

\noindent\textbf{Ablation of score functions.}\quad
To analyze the versatility of our methods, we further apply OAL to MSP \cite{hendrycks2017a} and GEN \cite{liu2023gen} (the softmax-based), KNN \cite{sun2022out} (the distance-based) on the CIFAR-100 benchmark. As shown in Table \ref{tab-score-ablation}, OAL works well for GEN \cite{liu2023gen} score and EBO \cite{liu2020energy} score. For MSP and KNN, we observe that the mean FPR@95 and AUROC decrease after using OAL for training the network. Therefore, our experiment results imply that OAL may has a positive impact on the softmax-based and the energy-based post hoc methods and has negative impacts on the distance-based approaches.


\begin{table}[t]
    \centering
    \caption{Ablation of the score function in our training framework on CIFAR-100 benchmark using ResNet-18 as the backbone. 
    }
    \label{tab-score-ablation}
    \resizebox{0.31\textwidth}{!}{
    \begin{tabular}{@{}cccc@{}}
    \toprule[2.5pt]
        \multirow{2}{*}{\textbf{Methods}} & \multicolumn{2}{c}{\textbf{Average}} & \multirow{2}{*}{\textbf{ID ACC}} \\ 
        ~& \textbf{FPR@95 $\downarrow$} & \textbf{AUROC $\uparrow$} &~ \\ 
        \midrule[1.5pt]
        MSP \cite{hendrycks2017a} & \textbf{57.89} & \textbf{79.19} & 76.22 \\  
        + OAL  & 60.02$\pm$3.93 & 79.10$\pm$1.47 & \textbf{76.89$\pm$0.20} \\ 
        \midrule[1.5pt]
        EBO \cite{liu2020energy} & 55.83  & 80.85 & 76.22 \\  
        + OAL & \textbf{51.30$\pm$4.14}  & \textbf{83.36$\pm$1.43} & \textbf{76.89$\pm$0.20} \\  
         \midrule[1.5pt]
        KNN \cite{sun2022out} &  \textbf{52.35} & \textbf{83.67} & 76.22 \\  
        + OAL & 53.01$\pm$2.75 & 83.39$\pm$0.40 & \textbf{76.89$\pm$0.20} \\ 
        \midrule[1.5pt]
        GEN \cite{liu2023gen} & 55.84 & 80.84 & 76.22 \\ 
        + OAL & \textbf{51.36$\pm$4.16} & \textbf{83.34$\pm$1.43} & \textbf{76.89$\pm$0.20} \\   



        \bottomrule[2.5pt]
    \end{tabular}
    }
\end{table}

\subsection{Score Distribution Visualization}
In this section, we aim to explore the effect of OAL training on the score distribution using EBO score. To this end, we visualize the score distribution of EBO using CIFAR-100 as the in-distribution data and Texture as the OOD data under the vanilla training, the VOS training, and OAL training, respectively. As shown in Figure \ref{ebo-score-distribution}, OAL reduces the overlap areas between the ID and OOD score distributions more than VOS does on the Texture dataset, preventing EBO from being overconfident of OOD data. This indicates that OAL has a stronger ability to enhance EBO’s OOD detection performance than VOS. Additionally, we observe that the standard deviation of the estimated scores on both ID and OOD data is reduced considerably after applying OAL, highlighting another advantage of our method, i.e., OAL enables EBO to produce more accurate OOD scores for test samples.



%
%

\section{Conclusion}
In this work, we present a novel outlier synthesis and training framework, named as OAL, facilitating outlier exposure training by synthesizing OOD data from the latent space of Stable Diffusion and developing a mutual information-based contrastive learning module to effectively regularize the model's decision boundary. Moreover, we develop knowledge distillation blocks to prevent degradation in ID classification accuracy while simultaneously enhancing the model's OOD performance. The experiments on the CIFAR-10 and CIFAR-100 benchmarks show that OAL outperforms many state-of-the-art post hoc methods and outlier synthesis-based methods, demonstrating the effectiveness of our framework.

\section*{Acknowledgments}
We thank Dr. Ke Fan, Zhuolin He and Shoumeng Qiu for helpful discussions.


\bibliographystyle{ACM-Reference-Format}
\bibliography{acmart}










\end{document}


\title{Enhancing OOD Detection Using Latent Diffusion (Supplementary Materials)}

\author{Heng Gao}
\email{hgao22@m.fudan.edu.cn}
\affiliation{%
  \institution{Fudan University}
  \city{Shanghai}
  \country{China}
}

\author{Jun Li}
\authornote{Corresponding author.}
\email{jun_li@fudan.edu.cn}
\affiliation{%
  \institution{Fudan University}
  \city{Shanghai}
  \country{China}
}

\ccsdesc[500]{Computing methodologies~Artificial intelligence}
\maketitle

\section{Visualization of the Generated Outliers}
\begin{figure}[htbp]
\renewcommand{\thefigure}{A}
\centerline{\includegraphics[width=0.56\linewidth]{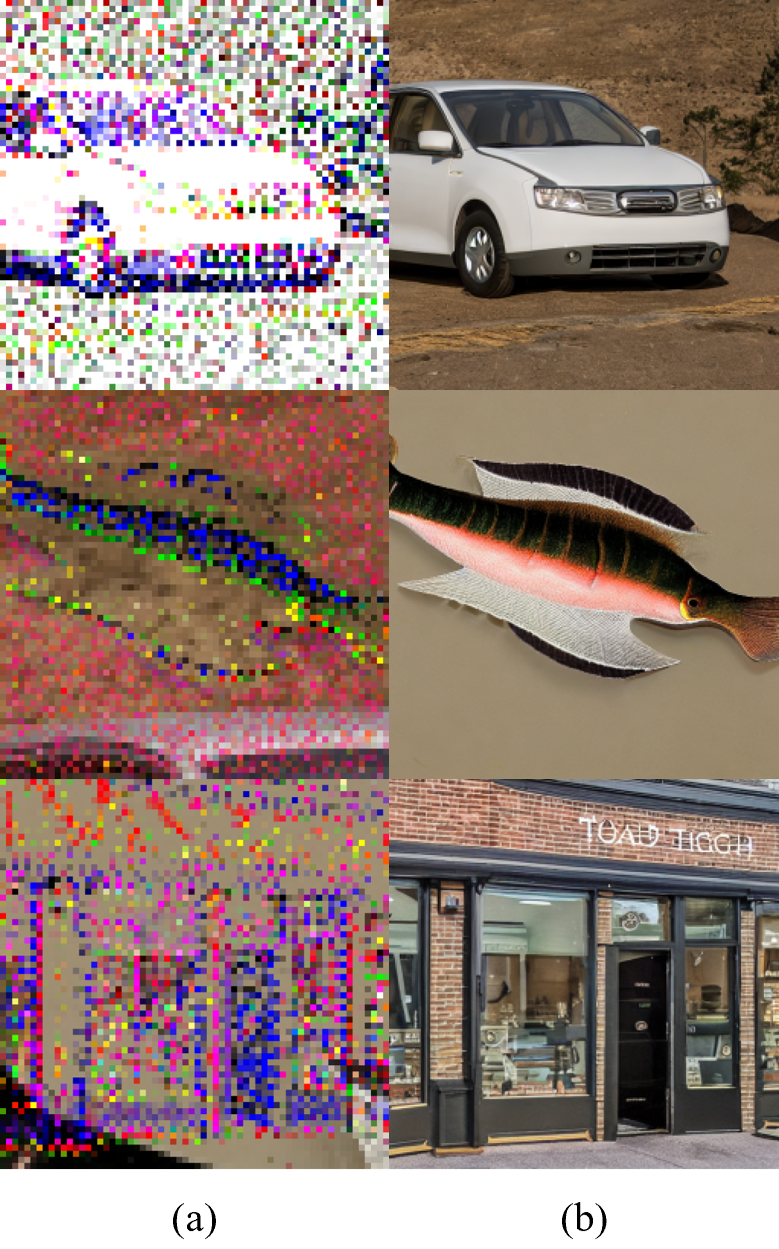}}
\caption{The visualization of the generated outliers using CIFAR-100 as the ID data. Column (a) represents the latent representations before taking $\text{Mean}(\cdot)$ operation; column (b) denotes the OOD images decoded from the latent representations in column (a).
}
\Description{1}
\label{generated-outliers}
\end{figure}

In this section, to better understand the outlier synthesis process of the diffusion model \cite{rombach2022high}, we visualize the generated outliers before taking $\text{Mean}(\cdot)$ operation on the first two dimensions of the latent representation and their corresponding images in the pixel space. The results are presented in Figure \ref{generated-outliers}.

\section{Inference Speed Comparison}
In this section, we report the inference speed of the transformer-based backbone and ResNet-18-based backbone on CIFAR-100 benchmark \cite{krizhevsky2009learning} using EBO score. As shown in Table \ref{tab-inference-speed}, ResNet-based backbone requires much less inference time than the transformer-based one.

\begin{table}[htb]
    \renewcommand{\thetable}{A}
    \centering
    \caption{The inference speed comparisons between transformer-based backbone and ResNet-18-based backbone on CIFAR-100 benchmark using EBO score. The input resolution are all set to $224\times 224$. Here, \textbf{Total Times} represents total inference time on six OOD datasets in CIFAR-100 benchmark.
    }
    \label{tab-inference-speed}
    \resizebox{0.3\textwidth}{!}{
    \begin{tabular}{ccc}
    \toprule[2.5pt]
        \textbf{Methods} & \textbf{Total Time $\downarrow$ (s)} & \textbf{GFLOPs $\downarrow$}\\ 
        \midrule[1.5pt]
        ResNet-18 & \textbf{\text{0.2}}$\times \textbf{\text{10}}^\textbf{\text{3}}$ & \textbf{1.82} \\ 
        ViT-B/16 & $1.98\times 10^3$ & 16.86 \\ 
        \bottomrule[2.5pt]
    \end{tabular}
    }
\end{table}

\bibliographystyle{ACM-Reference-Format}
\bibliography{acmart}








